\pdfoutput=1
\documentclass[11pt]{article}
\usepackage[table]{xcolor}
\usepackage[preprint]{acl}
\usepackage{amsmath}
\usepackage{amssymb}
\usepackage{booktabs}
\usepackage{multirow}

\usepackage{algorithm} 
\usepackage{algorithmicx}
\usepackage{algpseudocode}
\usepackage{subfigure}
\usepackage{times}
\usepackage{latexsym}
\usepackage{listings}
\usepackage{titlesec}
\usepackage[T1]{fontenc}
\usepackage[utf8]{inputenc}

\usepackage{microtype}

\usepackage{inconsolata}

\usepackage{graphicx}

\title{Attention with Dependency Parsing Augmentation\\for Fine-Grained Attribution}

\author{
  \textbf{Qiang Ding\textsuperscript{1,2}},
  \textbf{Lvzhou Luo\textsuperscript{1,2}},
  \textbf{Yixuan Cao\textsuperscript{1,2}},
  \textbf{Ping Luo\textsuperscript{1,2,3}},
\\
  \textsuperscript{1}Key Lab of Intelligent Information Processing of Chinese Academy of Sciences (CAS), 
  \\Institute of Computing Technology, CAS, Beijing 100190, China,
\\
  \textsuperscript{2}University of Chinese Academy of Sciences, CAS, Beijing 100049, China,
\\
  \textsuperscript{3}Peng Cheng Laboratory, Shenzhen 518066, China,
\\
  \small{
    \textbf{Correspondence:} \href{cauyixuan@ict.ac.cn}{caoyixuan@ict.ac.cn}, \href{luop@ict.ac.cn}{luop@ict.ac.cn}
  }
}

\begin{document}
\maketitle
\begin{abstract}
%Retrieval-augmented generation (RAG)
%is widely used in LLM-based QA systems across various domains. 
%However, it
%can produce statements that are not faithful to the retrieved documents, requiring meticulous verification. 
To assist humans in efficiently validating RAG-generated content, developing a fine-grained attribution mechanism that provides supporting evidence from retrieved documents for every answer span is essential.
Existing fine-grained attribution methods rely on model-internal similarity metrics between responses and documents, such as saliency scores and hidden state similarity.
However, these approaches suffer from either high computational complexity or coarse-grained representations.
Additionally, a common problem shared by the previous works is their reliance on decoder-only Transformers, limiting their ability to incorporate contextual information after the target span.
To address the above problems, we propose two techniques applicable to all model-internals-based methods.
First, we aggregate token-wise evidence through set union operations, preserving the granularity of representations.
Second, we enhance the attributor by integrating dependency parsing to enrich the semantic completeness of target spans.
For practical implementation, our approach employs attention weights as the similarity metric.
Experimental results demonstrate that the proposed method consistently outperforms all prior works.
%To address the high GPU memory consumption associated with calculating attention weights, we introduce a new routine that optimizes memory usage and accelerates the attribution process.
%Empirical results confirm that our attributor is significantly faster than prior methods.

\end{abstract}

\section{Introduction}

\begin{figure}[t]
    \centering
    \includegraphics[width=\linewidth]{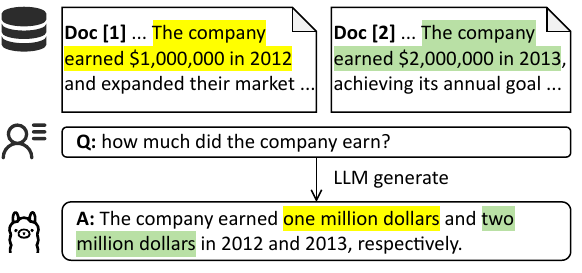}
    \caption{An example of fine-grained attribution, i.e., finding evidence from the retrieved documents for arbitrary target spans. Each highlighted span in the answer is a target, with evidence in the same background color. }
    \label{fig:fine-grained-attribution-example}
\end{figure}

Retrieval-augmented generation (RAG) enhances the factual recall of LLMs and has been applied in knowledge-intensive NLP tasks such as open-domain question answering \citep{lewis2020retrieval}.
However, the generated content may still deviate from the retrieved documents \citep{niu2024ragtruth}, necessitating careful verification, especially when used in safety-critical domains like finance.
To assist users in validating LLM-generated responses, QA systems must provide supporting evidence, also referred to as attribution or citation \citep{li2023survey}.
Furthermore, developing a fine-grained attribution mechanism that supplies evidence for arbitrary answer spans (as illustrated in Fig.~\ref{fig:fine-grained-attribution-example}) is particularly beneficial, as it allows users to efficiently verify the accuracy of individual segments within complex, long-form answers.
%This is the focus of this paper because it is inconvenient for users to check long and complex evidence for a complicated sentence or an entire response.
% , especially when one piece of evidence is a document. 
% Furthermore, the attributed pieces of evidence are usually documents, which is difficult for users to check.
% For example, to attribute the sentence "LeBron James earns most in NBA, with a salary of xxx million dollars", we may obtain two pieces of evidence: one states that LeBron James earns most in NBA; another one depicts the salary of LeBron James. 
%In contrast, fine-grained spans within sentences can have less but clearer evidence, which is more user-friendly. 
% Additionally, it is more favorable if the evidence is sentences rather than documents. 
% Therefore, we investigate fine-grained attribution in this paper.

Fine-grained attribution has been addressed by only two model-internals-based approaches: CCI \citep{yin2022interpreting, sarti2023inseq, qi2024model} and \textsc{HSSAvg} (Average Hidden State Similarity method, \citealp{phukan2024peering}).
CCI leverages saliency scores for attribution but requires gradient back-propagation for each target token, resulting in high computational complexity.
\textsc{HSSAvg} operates by (1) measuring the similarity between the average hidden states of the token span and those of a sliding window with a fixed size $W$ over the documents, and (2) selecting the highest-scoring window as the evidence.
However, the averaging operation over the target span introduces coarse granularity, limiting the precision of its representations.
A further challenge common to both methods lies in their reliance on the internal representations of decoder-only Transformers, which lack access to tokens that appear after the target span, constraining their contextual understanding (see Sec.~\ref{sec:dependency-parsing-augmentation} for details).
%the full context is inaccessible. 
%"NCAS can stand for the National Civic Art Society, the Australian Government's National Coaching Accreditation Scheme, the National Clinical Assessment Service, or the National Carbon Accounting System."

Building on insights into the limitations of previous methods, we propose a novel, effective, and efficient model-internals-based approach.
This approach consists of two techniques applicable to all model-internals-based approaches.
The first technique involves aggregating token-wise evidence using set unions, addressing the coarse-granularity issue in averaging hidden states while eliminating the need to recompute these averages for each new target span (see Sec.~\ref{sec:basic-algorithm} for details).
The second technique incorporates dependency parsing to enhance the attributor by integrating related tokens into the attribution of the target token, as illustrated by Fig.~\ref{fig:method-illustration} (see Sec \ref{sec:dependency-parsing-augmentation} for details).
For practical implementation, we utilize attention weights as the similarity metric due to their faster computation compared to gradient back-propagation and superior empirical performance over hidden state similarity.
Our experiments demonstrate that the proposed method surpasses all baseline approaches and generalizes effectively to sentence-level attribution, highlighting its practical value.
% This is different from the aggregation in \citet{qi2024model} which aggregates sentence-level evidence.
% We choose attention weights rather than hidden state similarity because attention weights are natural intermediate products of Transformers, which do not require extra computational overload beyond the generation process.

%However, attention-based methods face several challenges in real-world applications.
We also address two practical challenges in implementing our method: (1) the inaccessibility of attention weights from black-box sources and (2) the high GPU memory consumption required for their computation.
To tackle the first challenge, we approximate attention weights using open-source LLMs. 
For the second, we apply engineering optimizations to enable more efficient attention weight calculations, achieving significantly faster performance than prior approaches.

In summary, our contributions are:

1. We propose a novel model-internals-based fine-grained attributor that aggregates token-wise evidence through set unions, addressing the coarse granularity induced by averaging hidden states.

2. We propose leveraging dependency parsing to enhance decoder-based attributors by enriching the semantics of the target span.

3. The proposed method utilizes LLM attention weights for attribution and incorporates an optimized routine for efficient computation.

4. Our method sets a new state-of-the-art in fine-grained attribution and demonstrates strong generalization to sentence-level attribution.

\section{Related Work}

This section reviews related work across four aspects: (1) attribution, (2) non-faithfulness detection, (3) fact verification, and (4) other works using attention weights.

\noindent\textbf{Attribution.} Attribution methods can be categorized into three classes: self-generated attribution (or self-citation) \citep{thoppilan2022lamda, nakano2021webgpt, menick2022teaching}, retrieval-based attribution \citep{gao2023rarr, slobodkin2024attribute, sancheti2024post}, and model-internals-based attribution \citep{qi2024model, wang2024contextcite, phukan2024peering}.
The self-generated approach prompts or fine-tunes the LLM to produce citations during answer generation, but it does not guarantee that each statement has adequate citations \citep{gao2023enabling}.
The retrieval-based approach provides sentence-level citations by retrieving relevant results from external documents, ensuring adequate citations for each sentence.
Model-internals-based attribution relies on similarity metrics such as saliency scores or hidden state similarities to align the response with the prompt, retrieving evidence from prompt tokens with the highest similarity.
Of these three approaches, the first two primarily target sentence-level attribution, while some methods from the third category explore fine-grained attribution \citep{phukan2024peering, qi2024model}, which has been discussed in the Introduction.

\noindent\textbf{Non-Faithfulness Detection.} Attribution plays a valuable role in detecting non-faithfulness \citep{he2022rethinking, niu2024ragtruth}, a type of hallucination detection \citep{wang2020asking, muhlgay2024generating} that assesses whether the generated text is inconsistent with the \textit{input documents}.
Many non-faithfulness detection methods are black-box \citep{feng2023factkb, yue2023automatic, bazaga2024unsupervised, mishra2024fine, zhang2024enhancing}, resembling traditional Natural Language Inference (NLI) models but operating with longer contexts and greater complexity.
In these methods, attribution can be integrated into the detection pipeline to break down the complex NLI task, enhancing interpretability by providing supporting evidence for each statement.
An alternative approach to hallucination detection involves the model-internals methods \citep{li2023inference, hu2024lrp4rag, chuang2024lookback}, which offer the potential to unify attribution and non-faithfulness within a single model.
However, these methods have so far overlooked attribution.
%By producing sentence-level evidence, attribution reduces non-faithfulness detection to a sentence-level Natural Language Inference (NLI) task where the evidence serves as the premise and the statement as the hypothesis, a task readily handled by NLI-trained BERT models.
%Among these works, only one uses attention weights averaged over the prompt and the response \citep{chuang2024lookback} but does not consider attribution.

\noindent\textbf{Fact Verification.} Fact verification integrates hallucination detection with attribution, requiring not only an assessment of whether a claim is hallucinated but also evidence that supports or contradicts the claim \citep{guo2022survey}. 
In this domain, retrieval-based attribution is commonly used \citep{min2023factscore}.
While early studies (before the LLM era) employed attention weights from GNNs or other small models for attribution \citep{yang2019xfake, liu2020fine, chen2022loren}, using LLM attention weights for attribution remains unexplored in fact verification literature.
%As summarized by \citet{guo2022survey}, the conventional workflow of fact verification is 1. evidence retrieval, 2. verdict prediction, and 3. justification production, where attribution can be used in the third step.
%Most fact verification methods use retrieval to collect evidence for each statement \citep{min2023factscore, }.
%Although \citet{min2023factscore} study fact verification for sub-sentence-level claims, the proposed method cannot attach evidence to arbitrarily selected target spans.

\noindent\textbf{Other Works Using Attention Weights.} In attribution, attention weights have been utilized in small models like BERT \citep{clark2019bert, kobayashi2020attention}.
However, with the advent of LLMs, their use for attribution remains unexplored due to high memory costs and the lack of support from popular inference frameworks and proprietary models. 
For hallucination mitigation, \citet{huang2024opera} introduces a method that retrospects LLM generation whenever a hallucination-related ``aggregate pattern'' appears in the attention weights.

\section{Method}
\label{sec:method}

\begin{figure*}[t]
    \centering
    \includegraphics[width=0.9\linewidth]{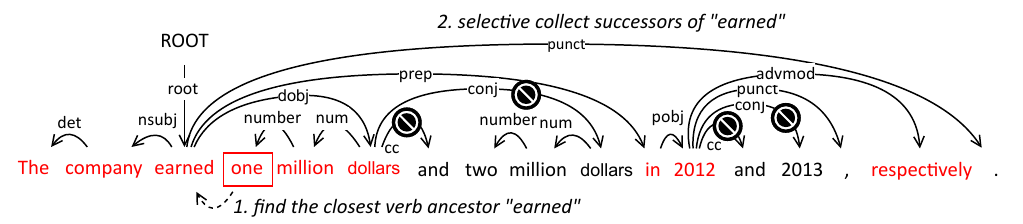}
    \caption{An illustration of dependency parsing augmentation. Suppose the target span is the token ``one''. The method first finds the closest verb ancestor of ``one'', i.e., ``earned'', and then collects successors of ``earned'', excluding unrelated coordinating constituents ``two million dollars'' and ``2013''. The resulting augmentation tokens are in red. Lastly, the attribution of ``one'' is updated by summing the attributions of the augmentation tokens.}
    \label{fig:method-illustration}
\end{figure*}

We begin by outlining the problem settings in Sec.~\ref{sec:problem-settings}, followed by a description of the attention-based method in Sec.~\ref{sec:basic-algorithm} and the dependency parsing augmentation in Sec.~\ref{sec:dependency-parsing-augmentation}.
Finally, we introduce our approach to addressing challenges in real-world applications in Sec.~\ref{sec:resolving-real-world-challenges}. 

\subsection{Problem Settings}
\label{sec:problem-settings}

Suppose an LLM generates a response $\mathbf{r}$ given the prompt composed of the retrieved documents $\mathbf{d}$ and the question $\mathbf{q}$.
%: $\mathbf{r} \sim p_{\text{LLM}}(\cdot|\mathbf{d}, \mathbf{q})$, where $p_{\text{LLM}}$ is the output probability of the LLM. 
The goal of \textit{fine-grained attribution} is, given an arbitrary target span $\mathbf{t} \subset \mathbf{r}$, to identify evidence from $\mathbf{d}$ that supports the \textit{atomic facts} (short statements that each contain one piece of information, \citealp{min2023factscore}) that involves $\mathbf{t}$.
An example is provided in Fig.~\ref{fig:fine-grained-attribution-example}, where for target span ``one million dollars'', the atomic fact is ``The company earned one million dollars in 2012'', and the evidence is ``The company earned \$1,0000,00 in 2012'' in the first document.

\subsection{The Basic Algorithm} 
\label{sec:basic-algorithm}

To solve fine-grained attribution, we propose attributing each target token in the span and then aggregating the attributions of tokens by the union of sets.
%using model-internals-based similarity metrics to 
Formally, let the documents, question, and response of the LLM be three sequences of tokens $\mathbf{d} = (d_1, d_2, ..., d_c)$, $\mathbf{q} = (q_1, q_2, ..., q_m)$, and $\mathbf{r} = (r_1, r_2, ..., r_n)$, respectively.
Assume the LLM can output a similarity metric $\mathbf{S} \in \mathbb{R}^{n\times(c+m)}$ between response tokens and prompt tokens (e.g., attention weights, hidden state similarity, and saliency scores).
Then the attribution scores are defined as\footnote{Here we assume the documents are located at the start of the prompt. If not, we need to replace $\mathbf{S}_{ij}$ with $\mathbf{S}_{i,j+o}$, where $o$ is the offset of the documents in the prompt.}
\begin{equation}
w(r_i, d_j) := \begin{cases}
\mathbf{S}_{i,j}, \text{if }\mathbf{S}_{i,j} \ge\text{top-}k(\mathbf{S}_i) \\
0, \text{otherwise}
\end{cases}
\label{eq:attribution-score}
\end{equation}
where $1\le i\le n$, $1\le j \le c$, top-$k(\mathbf{x})$ is the $k$-th largest component in vector $\mathbf{x}$, and $k$ is a hyper-parameter.
With these attribution scores, each response token $r_i$ is attributed to $e(r_i) := \{d_j|w(r_i, d_j) > 0\}$,
%, i.e., the intersection of the document and the top-$k$ prompt tokens scored by $\mathbf{W}_{i}\in \mathbb{R}^{c+m}$, where each evidence token $d_j\in e(r_i)$ is associated with an attribution score $w(r_i, d_j) := \mathbf{W}_{ij}$.
%For other document tokens, we define $w(r_j, d_j) := 0, \forall d_j \notin e(r_j)$.
and the aggregated evidence for the target span $\mathbf{t}$ is $e(\mathbf{t}) := \bigcup_{r_i \in \mathbf{t}} e(r_i) = \left\{d_j |\sum_{r_i\in\mathbf{t}} w(r_i, d_j) > 0\right\}$ with each token $d_j \in e(\mathbf{t})$ associated with an attribution score $w(\mathbf{t}, d_j) := \sum_{r_i\in \mathbf{t}} w(r_i, d_j)$. 
Furthermore, to avoid noisy and fragmented attribution\footnote{An example of noisy attribution is the widely-known attention sink \citep{kobayashi2020attention, xiao2024efficient, huang2024opera, zhang2024seeing} where a large amount of response tokens distribute a lot of attention weights to several isolated meaningless prompt tokens (e.g., BOS and punctuation marks), hindering the word alignment of attention weights}, we remove the isolated evidence tokens that are at least $\tau$ tokens away from other evidence tokens.
The pseudo-code of the whole algorithm is provided in Algorithm \ref{alg:basic-algorithm}.
In addition, the token-wise attributions $e(r_i)$ with scores $w(r_i, \cdot)$, $i=1,2, ..., n,$ can be reused by any target span in the same response, saving computational cost. 
We utilize this trick but omit it in the pseudo-code for brevity.

\subsection{Dependency Parsing Augmentation} 
\label{sec:dependency-parsing-augmentation}

We observe a common defect in previous works: the representations of decoder-only models cannot see the context that follows the target span, potentially missing relevant factual information that appears later. 
For instance, as shown in Fig.~\ref{fig:fine-grained-attribution-example}, decoder-only attributors cannot see ``million dollars'' or ``in 2012'' at the position of ``one'', increasing the difficulty of fine-grained attribution.
A straightforward solution is to allow the attributor to access subsequent context. 
However, extending the accessible scope to the entire context or full sentence is empirically shown to be ineffective in fully addressing the issue (see Sec. \ref{sec:ablating-dep}).
%We hypothesize that this is because the representations do not encode adequate information about the atomic facts that contain the target span.
%where two document snippets provide two different interpretations of ``NCAS,'' and the LLM chooses one of them at first and generates ``NCAS can stand for the National Centre for Atmospheric Science and it was founded in...''
%will struggle to determine which snippet ``NCAS'' refers to because the representation used for attributing ``NCAS'' cannot see the interpretation that follows.

To address this issue, we propose \textbf{DEpendency Parsing augmentation} (\textsc{Dep}).
This method first recognizes key elements --- such as subject, object, and predicate --- within atomic facts that contain the target span using dependency parsing.
For example, as illustrated in Fig.~\ref{fig:method-illustration}, given the target token ``one'', the method collects atomic fact elements as shown in red: ``The company earned one million dollars ... in 2012 ... respectively.'' 
It then leverages the attributions of these atomic fact elements to enhance the attribution of the target span.
The detailed algorithm proceeds as follows.

\noindent\textbf{1. Recognizing Atomic Fact Elements.} We assume that the elements of atomic facts can be extracted from the text.
For each token $r_i$, let these elements be $\mathcal{A}(r_i) \subset \mathbf{r}$.
This extraction is approximated by leveraging the dependency parse tree, constructed using the LAL-Parser \citep{mrini2020rethinking}, applied to the local sentence, as shown in Fig.~\ref{fig:method-illustration}. 
Since each node in this tree corresponds to a word while the attribution pertains to tokens, we need to align the two.
For simplicity, we temporarily assume they are the same, with alignment details provided in Appendix \ref{sec:word-token-alignment}.
Initially, we set $\mathcal{A}(r_i) \leftarrow \{r_i\}$.
Then, the algorithm appends $\mathcal{A}(r_i)$ with $r_i$'s closest verb ancestor $v$ and all $v$'s successors except punctuation marks.
If $\mathcal{A}(r_i)$ contains coordinating structures (e.g., tokens connected by ``and'' or ``or''), we use a rule-based subroutine (described in Appendix \ref{sec:excluding-coordinates}) to eliminate irrelevant coordinating constituents.
%retain the closest constituent to $r_i$ (based on traversal on the tree) along with its parallel constituents in other coordinating structures while discarding the remaining coordinating constituents.
For example, in Fig.~\ref{fig:method-illustration}, $\mathcal{A}(\text{``one''})$ includes the closest verb ancestor ``earned'' and all its successors except the irrelevant coordinating constituents ``two million dollars'' and ``2013.''

\noindent\textbf{2. Augmenting Attribution.} Using $\mathcal{A}(r_i)$, the attribution of token $r_i$ is augmented as $e(r_i) \leftarrow \bigcup_{a\in \mathcal{A}(r_i)} e(a)$, with attribution scores updated to $w(r_i, d_j) \leftarrow \sum_{a\in \mathcal{A}(r_i)} w(a, d_j)$.
The aggregate attribution is then computed based on the updated token-wise attribution, following the same approach as the basic algorithm.

\subsection{
Attention and its Computational Challenges
%Resolving the Challenges in Real-world Applications
}
\label{sec:resolving-real-world-challenges}

We choose attention weights as similarity scores due to their strong empirical performance.
The method is outlined as follows.
Let the attention weights between the response and the prompt at the $l$-th layer and the $h$-th head of the LLM be $\mathbf{A}^{(l, h)} \in\mathbb{R}^{n\times (c+m)}$, where the LLM has $L$ layers and $H$ heads.
Specifically, $\mathbf{A}^{(l,h)}_i$ is the attention weights used to predict $r_i$.
For a Transformer decoder, $\mathbf{A}^{(l, h)}_{ij}$ is the attention weight from $r_{i-1}$ to $d_{j}$.
For a Transformer encoder, $\mathbf{A}^{(l, h)}_{ij}$ is the attention weight from $r_i$ to $d_j$.
Based on these attention weights, the similarity score is computed by averaging the attention weights from a selected layer $L^*$ across all heads: $\mathbf{S} := \frac 1H \sum_{h=1}^H \mathbf{A}^{(L^*, h)}$.

Another potential similarity metric is cosine similarity among hidden states.
Let $\mathbf{h}^{\text{p}} \in \mathbb{R}^{(c+m)\times d}$ and $\mathbf{h}^{\text{r}} \in \mathbb{R}^{r\times d}$ denote the hidden states of the prompt and the response, respectively, where $d$ is the dimension of hidden states.
The hidden state cosine similarity is computed as: $\mathbf{S}_{ij} := \frac{\mathbf{h}_i^{\text{r}}\cdot\mathbf{h}_j^{\text{p}}}{\Vert \mathbf{h}_i^{\text{r}}\Vert\cdot\Vert \mathbf{h}_j^{\text{p}}\Vert}$.
We will empirically show that attention-based similarity is better than hidden state cosine similarity.

However, the attention-based approach has the following two challenges in real-world application. 

\noindent\textit{Inaccessible Attention Weights.} The attention weights are inaccessible under the following conditions: (1) the LLM is proprietary, such GPT-4 \citep{achiam2023gpt}; (2) the inference framework, such as vLLM \citep{kwon2023efficient}, does not support outputting attention weights; (3) the response may be generated by other black-box sources, e.g., humans. 
To address this, we leverage open-sourced LLMs to approximate the attention weights.
Empirical results demonstrate that these approximate attention weights offer reliable attribution.

\noindent\textit{High Memory Overload.}
%Suppose the attention weight calculation and generation processes co-occur. 
In the Huggingface \citep{wolf2020transformers} implementation\footnote{The implementation details are provided in Appendix \ref{sec:huggingface-implementation}.}, attention weights and the KV cache are stored simultaneously in GPU memory when calculating attention weights from the response to the prompt.
This dual memory load can quickly exhaust the GPU memory.
Therefore, we propose a specialized routine for calculating attention weights.
We calculate attention weights after the generation process.
The model processes the concatenation of the response and the prompt with FlashAttention \citep{dao2022flashattention} and NF4 quantization \citep{dettmers2023qlora}, exiting early  after the $(L^*-1)$-th layer without producing attention weights or storing the KV cache. 
From the resulting hidden states, we calculate the attention weights from the response to the prompt, yielding the desired similarity metric.
%This approach has been empirically validated to perform effectively and efficiently on an NVIDIA RTX 3090 GPU for a 7B-parameter LLM.
% as the recent memory-efficient self-attention implementations, e.g., FlashAttention \citep{dao2022flashattention}, avoid calculating the entire attention weight matrices to save GPU memory.
% Considering $G \ll P$ in practice, the above routine largely reduces the memory complexity.
\section{Experiments}

We begin by evaluating our methods on fine-grained attribution in Sec.~\ref{sec:eval-fine-grained-attribution}, with the corresponding ablation study presented in Sec.~\ref{sec:ablation-study}.
We then assess the faithfulness of our methods to the generators in Sec.~\ref{sec:eval-faithfulness}.
The evaluation of sentence-level attribution is in Sec.~\ref{sec:eval-sentence-level-attribution}.
Lastly, we compare the latency of all fine-grained attributors in Sec.~\ref{sec:eval-latency}.

\subsection{Evaluating Fine-Grained Attribution}
\label{sec:eval-fine-grained-attribution}

\noindent\textbf{Benchmark.} Following \citet{phukan2024peering}, we use two datasets QuoteSum \citep{schuster2024semqa} and VERI-GRAN \citep{phukan2024peering} to evaluate fine-grained attribution.
Each instance of these two datasets contains a question, retrieved passages, and an answer, where the question and the retrieved passages combine to form the prompt\footnote{We use the same prompt template as \citet{phukan2024peering}.}.
The task is to identify the evidence passage (or evidence sentence in the case of VERI-GRAN) corresponding to each context-sensitive span in the answer, where the span is determined based on the similarity of LLM hidden states \citep{phukan2024peering}, and the evidence passage is labeled by human annotators. 
Additional details about the datasets are provided in Table~\ref{tab:statistics}.
The evaluation metric is the accuracy of the predicted evidence passage for these spans.
To apply our method and \textsc{CCI} to this benchmark, we predict the passage with the highest cumulative attribution score, i.e., $\arg\max_{\mathbf{d}_i}\sum_{d\in \mathbf{d}_i} w(\mathbf{t}, d),$ where $\mathbf{d}_i$ is the $i$-th passage and $\mathbf{t}$ is the target span.
For \textsc{HSSAvg}, we select the passage containing the highest-scoring sliding window as the prediction.

\noindent\textbf{Baselines.} We include the only two prior works on fine-grained attribution as our baselines, CCI \citep{sarti2023inseq, qi2024model}, also known as contrastive feature attribution \citep{yin2022interpreting}, and \textsc{HSSAvg} \citep{phukan2024peering}.
Additionally, we include the reported results of GPT-4 from \citet{phukan2024peering}.

\noindent\textbf{Models.} We conduct experiments using two models: Llama2 7B Chat\citep{touvron2023llama2} and Qwen2 7B Instruct \citep{yang2024qwen2}.
For \textsc{HSSAvg}, hidden states are extracted from the $\lfloor \frac{L}{2}\rfloor$-th layer, as \citet{phukan2024peering} report that this method performs well across different models when using hidden states from the middle layers.
For our methods, the attention weights are from the $(\lfloor \frac L 2 \rfloor+1)$-th layer --- just above the hidden states used for \textsc{HSSAvg} --- due to their strong empirical performance on validation sets.
To address GPU memory limitations, we apply NF4 quantization to CCI\footnote{Our methods also use NF4 quantization, as Sec.~\ref{sec:method} states.}.

\noindent\textbf{Hyperparameters.} We set the token-wise evidence size $k=2$ and the threshold for recognizing isolated tokens $\tau = 2$ for our method (experiments on sensitivity to hyperparameters are presented in Appendix \ref{sec:hyperparameter-sensitivity}).
For \textsc{HSSAvg}, we configure the window size $W = 8$, as this setting yields good performance on validation sets. 
For \textsc{CCI}, following the original paper, we choose the top-scoring three tokens as the evidence for each target token.

\noindent\textbf{Results.} The accuracy of fine-grained attribution is reported in Table~\ref{tab:experiment-1-results}, where we refer to our basic algorithm and its augmented version as \textsc{AttnUnion} and \textsc{AttnUnionDep}, respectively.
Across both datasets, \textsc{AttnUnion} achieves comparable performance to \textsc{HSSAvg}, while \textsc{AttnUnionDep} consistently outperforms \textsc{HSSAvg} and GPT-4, setting a new SOTA in fine-grained attribution and demonstrating the effectiveness of \textsc{Dep}. 
%\footnote{It is worth noting that the reproduced results of \textsc{HSSAvg} are much lower than those reported in the original paper. We implemented \textsc{HSSAvg} following its paper's guidelines and tuned the hyperparameters on validation sets, but we were unable to replicate the reported results. }

% \usepackage{booktabs}
%\begin{table*}[t]
%\centering
%\begin{tabular}{@{}cccc@{}}
%\toprule
%\textbf{Model}           & \textbf{Method}       & \textbf{QuoteSum} & \textbf{VERI-GRAN} \\ \midrule
%GPT-4                    & 2-shot prompting      & 0.906    & 0.621     \\
%\hline
%\multirow{3}{*}{Qwen2}   & \textsc{HSSAvg}                & 0.804    & 0.671     \\
%                         & AttnUnion (Ours)      & 0.860    & 0.652     \\
%                         & AttnUnion+dep (Ours)  & \underline{0.933} & \textbf{0.846}     \\
%\hline
%\multirow{4}{*}{Llama2}  & \textsc{HSSAvg} \citep{phukan2024peering} & 0.875    & 0.773     \\
%                         & \textsc{HSSAvg}                & 0.771    & 0.645     \\
%                         & AttnUnion (Ours)      & 0.867    & 0.650     \\
%                         & AttnUnion+dep (Ours)  & \textbf{0.941} & \underline{0.791}     \\ 
%\bottomrule
%\end{tabular}
%\caption{}
%\label{tab:experiment-1-results}
%\end{table*}

% \usepackage{booktabs}
% \usepackage{multirow}
\begin{table}[t]
\centering
\resizebox{\linewidth}{!}{
\begin{tabular}{@{}lccc@{}}
\toprule
& \textbf{Model} & \textbf{QuoteSum} & \textbf{VERI-GRAN} \\ 
\midrule
\rowcolor{gray!20}\multicolumn{4}{c}{\textbf{Baselines}} \\
\textsc{2-shot}             & GPT-4$^\dagger$ & 90.6     & 62.1     \\ 
\cmidrule(r){2-4}
\multirow{3}{*}{\textsc{HSSAvg}} & Qwen2 & 80.4    & 67.1     \\
                                 & Llama2 & 77.1    & 64.5     \\
                                 & Llama2$^{\dagger} $ & 87.5 & 77.3    \\ 
\cmidrule(r){2-4}
\multirow{2}{*}{\textsc{CCI}} & Qwen2 & 71.3 & 64.5 \\
                              & Llama2 & 72.2 & 59.0 \\
%\hline
\rowcolor{gray!20}\multicolumn{4}{c}{\textbf{Our Methods}} \\
\multirow{2}{*}{\textsc{AttnUnion}} & Qwen2  & 79.4    & 70.9     \\
                                    & Llama2 & 81.3    & 66.7     \\ 
\cmidrule(r){2-4}
\multirow{2}{*}{\textsc{AttnUnionDep}} & Qwen2  & \underline{93.3} & \textbf{84.6} \\ 
                                       & Llama2 & \textbf{94.0}    & \underline{78.2} \\ \bottomrule
\end{tabular}
}
\caption{Accuracy (\%) of fine-grained attribution on QuoteSum and VERI-GRAN. The best and the second-best entries are marked in bold and underlined, respectively. Cited results are marked by $\dagger$.}
\label{tab:experiment-1-results}
\end{table}

% \usepackage{booktabs}
% \usepackage{multirow}
\begin{table}[t]
\resizebox{\linewidth}{!}{
%\begin{tabular}{@{}cccc@{}}
%\toprule
%Model                     & Method           & QuoteSum        & VERI-GRAN       \\ 
%\midrule
%\multirow{3}{*}{Qwen 2}   & \textsc{HSSAvgDep}    & 79.3           & 62.8           \\
%                          & HSSUnionDep  & 88.3           & 761           \\
%                          & AttnCompSent & 93.6 (-0.1) & 79.9 (-4.7) \\
%\cmidrule(r){2-4}
%\multirow{3}{*}{Llama 2}  & \textsc{HSSAvgDep}    & 72.2           & 60.7           \\
%                          & HSSUnionDep  & 89.2           & 75.6           \\
%                          & AttnCompSent & 93.2 (-0.9) & 76.1 (-3.0) \\
%\bottomrule
%\end{tabular}
\begin{tabular}{@{}lccc@{}}
\toprule
& \textbf{Model} & \textbf{QuoteSum} & \textbf{VERI-GRAN} \\ 
\midrule
%\rowcolor{gray!20}\multicolumn{4}{c}{Qwen2} \\
%\textsc{AttnUnion}  & & \\
%+ \textsc{Dep}      & 93.3         & 84.6 \\ 
%+ \textsc{SentComp} & 92.5 (-0.8)  & 79.9 (-4.7) \\
%+ \textsc{JinaBERT} & 26.5 (-66.8) & 6.0 (-78.6) \\
%\hline
%\textsc{HSSUnion}   & & \\
%+ \textsc{Dep}      & 88.3         & 76.1 \\
%+ \textsc{JinaBERT} & 72.7 (-15.6) & 58.5 (-17.6) \\

%\rowcolor{gray!20}\multicolumn{4}{c}{Llama2} \\
%\textsc{AttnUnion}  & & \\
%+ \textsc{Dep}      & 94.1         & 79.1 \\
%+ \textsc{SentComp} & 93.2 (-0.9)  & 76.1 (-3.0) \\
%+ \textsc{JinaBERT} & 26.5 (-67.6) & 6.0 (-73.1) \\
%\hline
%\textsc{HSSUnion}   & & \\
%+ \textsc{Dep}      & 89.2         & 75.6 \\
%+ \textsc{JinaBERT} & 72.7 (-16.5) & 58.5 (-17.1) \\

%\rowcolor{gray!20}\multicolumn{4}{c}{\textsc{Union} vs. \textsc{Avg}} \\
\rowcolor{gray!20}\multicolumn{4}{c}{\textbf{\textsc{Union} vs. \textsc{Avg} w/o \textsc{Dep}}} \\
\multirow{2}{*}{\textsc{HSSUnion}} & Qwen2  & 80.0 & 73.1 \\
                                   & Llama2 & 80.1 & 65.3 \\
\cmidrule{2-4}
\multirow{2}{*}{\textsc{HSSAvg}}   & Qwen2 & 80.4    & 67.1     \\
                                 & Llama2 & 77.1    & 64.5     \\
\rowcolor{gray!20}\multicolumn{4}{c}{\textbf{\textsc{Union} vs. \textsc{Avg} w/ \textsc{Dep}}} \\
\multirow{2}{*}{\textsc{HSSUnionDep}} & Qwen2 & 89.4 & 81.9    \\
                                    & Llama2 & 88.6  & 80.3     \\
\cmidrule{2-4}
\multirow{2}{*}{\textsc{HSSAvgDep}} & Qwen2 & 79.3 & 62.8     \\
                                    & Llama2 & 72.2 & 60.7    \\
\bottomrule
\end{tabular}
}
\caption{Accuracy (\%) of \textsc{Union} vs. \textsc{Avg} on QuoteSum and VERI-GRAN. 
%We copy the results of \textsc{AttnUnionDep} here for clarity of comparison. The gap between the ablated method and \textsc{AttnUnionDep} with the same model is shown in the braces.
}
\label{tab:union-vs-avg}
\end{table}

\subsection{Ablation Study} 
\label{sec:ablation-study}

\begin{figure*}[t]
    \centering
    \setlength{\abovecaptionskip}{0.cm}
    \subfigure[Qwen2 7B]{
        \includegraphics[width=0.48\linewidth]{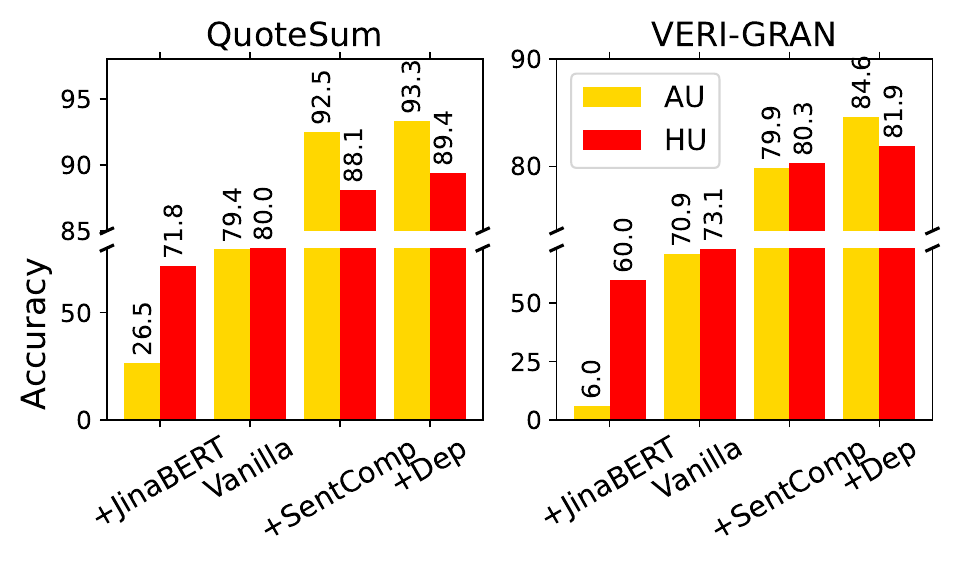}
    }
    \subfigure[Llama2 7B]{
        \includegraphics[width=0.48\linewidth]{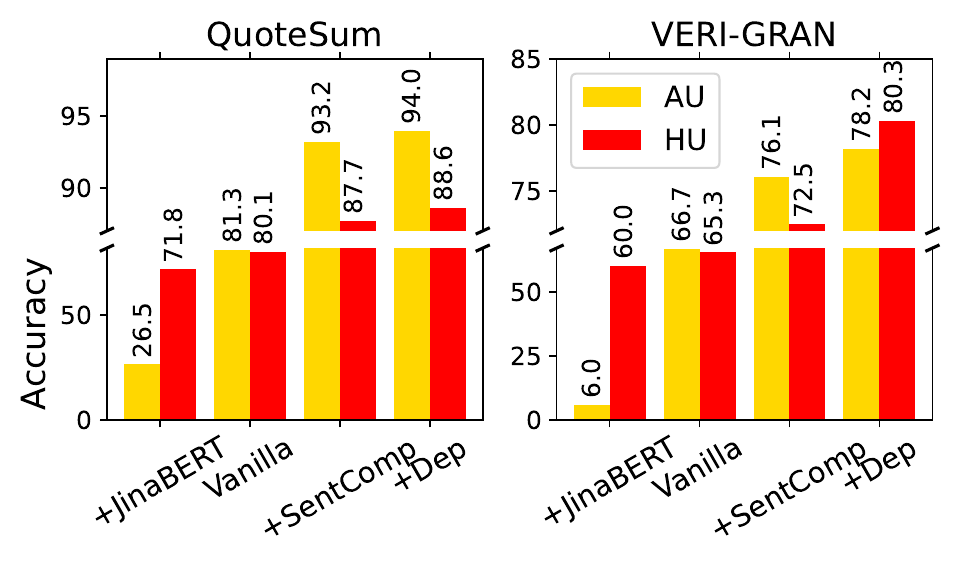}
    }
    \caption{Results of ablating \textsc{Dep}. Here AU and HU represent \textsc{AttnUnion} and \textsc{HSSUnion}, respectively.}
    \label{fig:ablating-dep}
\end{figure*}

In this section, we conduct ablation studies about \textsc{Dep}, \textsc{Attn}, and \textsc{Union}.
%We do not consider \textsc{CCI} due to its high computational complexity.

\subsubsection{Ablating \textsc{Dep}}
\label{sec:ablating-dep}

\noindent\textbf{Research Question.} Previous results have confirmed the effectiveness of \textsc{Dep}. 
However, an important question remains: \textit{Are there any simpler alternatives to \textsc{Dep} that allow the model representations to see the context following the target span?}

\noindent\textbf{Experiment Setting.} We empirically compare two alternatives with \textsc{Dep}\footnote{Another more complicated alternative of \textsc{Dep} is discussed in Appendix \ref{sec:additional-dep-alternative}, which is empirically not as good as \textsc{Dep}.}: 1. extending the span to the entire local sentence (\textsc{SentComp}); 2. leveraging representations from bidirectional attention models, such as BERT, for attribution.
The benchmarks, backbones, and hyper-parameters remain consistent with those used in the previous experiment.

Specifically, the second alternative leverages JinaBERT's \citep{gunther2023jina} attention weights or hidden state similarity (HSS) as the similarity metric.
We refer to the HSS-based method as \textsc{HSSUnion}.
The attention weights are from the 5th layer, while the hidden state similarities are from the final layer, as these configurations demonstrated strong performance on the validation sets.

\noindent\textbf{Results \& Insights.} The results are shown in Fig.~\ref{fig:ablating-dep}, demonstrating that all alternatives consistently underperform compared to \textsc{Dep}.
The results indicate that recognizing atomic facts via \textsc{Dep} is more effective than extending the accessible context to the entire context or full sentence.

\subsubsection{\textsc{Attn} vs. \textsc{HSS}}

\noindent\textbf{Research Question.} While calculating attention weights is significantly faster than gradient back-propagation, it incurs a similar computational cost to computing hidden state similarity. 
This raises the question: \textit{Is using attention weights more effective than hidden state similarity?}

%We verify the necessity of other components of the proposed method, i.e., using attention weights as attribution scores and aggregating token-wise attribution by the union of sets.

\noindent\textbf{Results.} We analyze the results from Fig.~\ref{fig:ablating-dep} by comparing adjacent bars of \textsc{AttnUnion} and \textsc{HSSUnion}.
Excluding the underperforming \textsc{JinaBERT} and \textsc{Vanilla} settings, \textsc{AttnUnion} outperforms \textsc{HSSUnion} in most cases, highlighting the effectiveness of using attention weights as attribution scores.

\subsubsection{\textsc{Union} vs. \textsc{Avg}}

\noindent\textbf{Research Question.} The final component to examine is the aggregation by union.
We pose the question: \textit{Is \textsc{Union} more effective than \textsc{Avg}?}

\noindent\textbf{Experiment Setting.} We compare \textsc{Union} and \textsc{Avg} both with and without the integration of \textsc{Dep}.
To incorporate \textsc{Dep} with \textsc{HSSAvg}, we first expand the target span using \textsc{Dep}, i.e., $\mathbf{t}\leftarrow \bigcup_{r_i\in \mathbf{t}}\mathcal{A}(r_i)$, and then apply \textsc{HSSAvg} to the new span. 
All other experimental settings remain consistent with those used in the previous experiment.

\noindent\textbf{Results.} The results, presented in Table~\ref{tab:union-vs-avg}, show that while \textsc{Union} underperforms \textsc{Avg} without \textsc{Dep}, it surpasses \textsc{Avg} by a significant margin --- at least 9.0 percent points --- when \textsc{Dep} is applied.
%which shows the necessity of aggregating by the union when \textsc{Dep} is present.

\noindent\textbf{Insights.} The results suggest that \textsc{Union} is most effective when used in conjunction with \textsc{Dep}. 
Conversely, combining the results of Table~\ref{tab:experiment-1-results} and Table~\ref{tab:union-vs-avg}, we conclude that \textsc{Dep} also works best with \textsc{Union}, as \textsc{HSSAvgDep} performs worse than \textsc{HSSAvg}.
This outcome is reasonable since the expanded target spans in \textsc{HSSAvgDep} introduce more context semantics, resulting in averaged hidden states that dilute fine-grained information.

\subsection{Faithfulness of fine-grained Attribution}
\label{sec:eval-faithfulness}

% \usepackage{booktabs}
% \usepackage{multirow}
% \usepackage[normalem]{ulem}
\begin{table}[t]
\centering
\resizebox{\linewidth}{!}{
\begin{tabular}{@{}lccccc@{}}
\toprule
& \textbf{Model} & \multicolumn{2}{c}{\textbf{QuoteSum}} & \multicolumn{2}{c}{\textbf{VERI-GRAN}} \\ 
                              &      & \texttt{Qw.} & \texttt{Ll.} & \texttt{Qw.} & \texttt{Ll.} \\
\midrule
\rowcolor{gray!20}\multicolumn{6}{c}{\textbf{Baselines}} \\
\textsc{Random}               &       & 0.81 & 1.09 & 0.13 & 0.30 \\
\cmidrule{2-6}
\multirow{2}{*}{\textsc{HSSAvg}}       & Qwen2 & 5.85 & 7.94 & 3.47 & 5.69 \\
                              & Llama2 & 5.97 & 8.05 & \textbf{3.50} & \underline{5.76} \\
\cmidrule{2-6}
\multirow{2}{*}{\textsc{HSSAvgDep}}    & Qwen2 & 5.77 & 7.74 & 3.26 & 5.29 \\
                              & Llama2 & 5.39 & 7.62 & 3.18 & 5.52 \\
\cmidrule{2-6}
\multirow{2}{*}{\textsc{CCI}} & Qwen2 & 5.70 & 7.39 & 3.47 & 5.39 \\
                              & Llama2 & 5.43 & 7.76 & 3.19 & 5.66 \\
\cmidrule{2-6}
\textsc{Oracle}               &        & 6.41 & 8.49 & 4.03 & 6.60 \\
%\midrule
\rowcolor{gray!20}\multicolumn{6}{c}{\textbf{Our Methods}} \\
\multirow{2}{*}{\textsc{AttnUnion}}    & Qwen2 & 4.08 & 5.53 & 2.39 & 4.40 \\
                              & Llama2 & 4.60 & 5.96 & 2.61 & 4.51 \\
\cmidrule{2-6}
\multirow{2}{*}{\textsc{AttnUnionDep}} & Qwen2 & \underline{6.17} & \underline{8.09} & \underline{3.48} & \textbf{5.79} \\
                              & Llama2 & \textbf{6.18} & \textbf{8.13} & 3.47 & \underline{5.76} \\
\bottomrule
\end{tabular}
}
\caption{Log probability drops on QuoteSum and VERI-GRAN with generator Qwen2 (\texttt{Qw.}) and Llama2 (\texttt{Ll.}). The best and the second best (except \textsc{Oracle}) entry is marked in bold and underlined, respectively.}
\label{tab:experiment-2-results}
\end{table}

\noindent\textbf{Research Question.} We have shown that the proposed method accurately identifies the human-labeled evidence passages, indicating a strong alignment between our approach and human annotations. 
However, it remains to be verified whether our method is faithful to the generation model --- specifically, \textit{does the attributed evidence directly influence the generator to produce the target span?}

\noindent\textbf{Experiment Setting.} To quantify the causal effect, we follow the approach of \citet{wang2024contextcite}, which involves removing the evidence from the prompt, rerunning the generator, and measuring the log predictive probability drop of the target span before and after the removal.
First, We use Llama2 or Qwen2 to generate answers with greedy decoding on QuoteSum and VERI-GRAN.
Next, in each generated answer, we apply CTI \citep{qi2024model} to identify context-sensitive tokens, which serve as target spans.
We then use the attributor to locate evidence for each target span.
Finally, we calculate the log probability drops as follows.
Let the documents be $\mathbf{d}_1, ..., \mathbf{d}_C$ with the attributed document $\mathbf{d}_i$.
The \textit{log probability drop} is 
$$
\log \frac{p_{\text{LLM}}(\mathbf{t}|\mathbf{d}_{1}, ..., \mathbf{d}_{C}, \mathbf{q}, \mathbf{r}_{\prec \mathbf{t}})} {p_{\text{LLM}}(\mathbf{t}|\mathbf{d}_{1}, ..., \mathbf{d}_{i-1}, \mathbf{d}_{i+1}, ..., \mathbf{d}_{C}, \mathbf{q}, \mathbf{r}_{\prec \mathbf{t}})},
$$
where $p_{\text{LLM}}$ is the output probability of the LLM, $\mathbf{t}$ is the target span, and $\mathbf{r}_{\prec \mathbf{t}}$ is the response prefix preceding $\mathbf{t}$.
A larger log probability drop indicates greater faithfulness of the attribution to the generation process.

%\noindent\textbf{Baselines.}
We include \textsc{HSSAvg}, \textsc{HSSAvgDep}, and \textsc{CCI} as baselines.
Additionally, we introduce two extra baselines, \textsc{Random} and \textsc{Oracle}.
\textsc{Random} randomly selects an evidence passage, with the experiment repeated three times to calculate the average performance.
\textsc{Oracle}, on the other hand, performs a brute-force search to identify the passage that results in the highest log-probability drop, using that passage as the evidence.

\noindent\textbf{Results \& Insights.} The results in Table~\ref{tab:experiment-2-results} show that \textsc{Dep} significantly enhances the faithfulness of \textsc{AttnUnion}.
In most cases, \textsc{AttnUnionDep} outperforms baselines except \textsc{Oracle} and closely approaches its performance, indicating that \textsc{AttnUnionDep} offers greater faithfulness than previous methods.
Interestingly, all methods maintain their faithfulness even when using different models from the generator, showing their flexibility and independence from the generator's backbone.

\subsection{Sentence-level Attribution}
\label{sec:eval-sentence-level-attribution}

% \usepackage{booktabs}
% \usepackage{multirow}
% \usepackage[normalem]{ulem}
\begin{table}[t]
\centering
\resizebox{\linewidth}{!}{
\begin{tabular}{@{}lcccccc@{}}
\toprule
&  \multicolumn{3}{c}{\textbf{ELI5}} & \multicolumn{3}{c}{\textbf{ASQA}} \\
\cmidrule(r){2-4} \cmidrule(r){5-7}
%& \textbf{Fluency} & \textbf{Correctness} & \multicolumn{2}{c}{\textbf{Citation}} & \textbf{Fluency} & \textbf{Correctness} & \multicolumn{2}{c}{\textbf{Citation}} \\
%\cmidrule(r){2-2} \cmidrule(r){3-3} \cmidrule(r){4-5} \cmidrule(r){6-6} \cmidrule(r){7-7} \cmidrule(r){8-9} 
& \textbf{R} & \textbf{P} & \textbf{F1} & \textbf{R} & \textbf{P} & \textbf{F1}\\
\midrule

\rowcolor{gray!20}\multicolumn{7}{c}{\textbf{Qwen2}} \\
\textsc{SelfCitation}  & 31.1 & 30.1 & 30.6 & 57.4 & 53.3 & 55.3 \\
+ \textsc{HSSAvg} & 24.5 & 22.4 & 23.4 & 35.0 & 25.9 & 29.8 \\
+ \textsc{CCI}    & 35.7 & 18.9 & 24.7 & 59.4 & 33.6 & 42.9 \\
+ \textsc{AttnUnion} & 39.4 & 29.3 & 33.6 & 65.7 & 49.0 & 56.1 \\
\cmidrule{2-7}
\textsc{AttrFirst} & \underline{69.3} & \textbf{69.1} & \underline{69.2} & \underline{63.4} & \textbf{69.9} & \underline{66.5} \\
+ \textsc{AttnUnion} & \textbf{81.9} & \underline{64.8} & \textbf{72.3} & \textbf{87.2} & \underline{60.7} & \textbf{71.6} \\

\rowcolor{gray!20}\multicolumn{7}{c}{\textbf{Llama2}} \\
\textsc{SelfCitation}  & 20.1 & 15.3 & 17.4 & \underline{55.4} & \underline{51.9} & \underline{53.6} \\
+ \textsc{HSSAvg} & 18.0 & 16.1 & 17.0 & 33.7 & 22.6 & 27.1 \\
+ \textsc{CCI} & \underline{26.2}$^\dagger$ & \textbf{29.1}$^\dagger$ & \textbf{27.6}$^\dagger$ & 42.3 & 22.6 & 29.5 \\
+ \textsc{AttnUnion} & \textbf{28.1} & \underline{26.9} & \underline{27.5} & \textbf{62.1} & \textbf{52.6} & \textbf{57.0} \\
\bottomrule
\end{tabular}
}
\caption{The citation quality of sentence-level attribution (\textbf{R} and \textbf{P} are citation recall and precision, respectively). ``+X'' means using X to attribute the generation of the \textsc{SelfCitation}/\textsc{AttrFirst} with the original citations removed. Cited results are marked by $\dagger$.}
\label{tab:sentence-level-attribution}
\end{table}

Fine-grained attribution can be easily applied to sentence-level attribution, as long as we select sentences as the target spans.
In the following, we evaluate the performance of sentence-level attribution of \textsc{AttnUnion} on commonly used benchmarks and compare it with the current SOTA\footnote{We do not use \textsc{AttnUnionDep} because \textsc{Dep} does not change the output evidence of \textsc{AttnUnion} in this experiment.}.

\noindent\textbf{Benchmarks.} Following \citet{gao2023enabling}, we conduct experiments on datasets ASQA \citep{stelmakh2022asqa} and ELI5 \citep{fan2019eli5}.
In these datasets, each question is attached with 100 documents, and the top 5 documents are used as the retrieved documents (the vanilla setting of \citet{gao2023enabling}).
Following ALCE \citep{gao2023enabling}, we use metrics MAUVE \citep{pillutla2021mauve}, EM/claim recall, citation recall, and citation precision.
The first two measure the generation quality, and the last two measure the citation quality.

\noindent\textbf{Baselines.} The baselines include all previously introduced baselines and two baselines of sentence-level attribution, the vanilla version of self-citation \citep{gao2023enabling} and Attribute First Then Generate (\textsc{AttrFirst} for short, \citealp{slobodkin2024attribute}).
To rule out the effect of different prompts, following \citet{qi2024model}, all fine-grained baselines share the prompt and the generated response of self-citation (with the original citations removed) to attribute. 
To compare our method with \textsc{AttrFirst}, we separately evaluate our method on the generated results of \textsc{AttrFirst}.
%except \textsc{AttrFirst} (due to its complex generation routine)

\noindent\textbf{Backbones.} The backbones are by default Qwen2 7B and Llama2 7B. 
However, due to the context length required by \textsc{AttrFirst} exceeds the max context length of Llama2 7B, we only evaluate \textsc{AttrFirst} on Qwen2 7B.

\noindent\textbf{Hyperparameters.} For fine-grained methods, because the benchmarks allow outputting multiple citations for a target sentence, \textsc{AttnUnion}, \textsc{HSSAvg}, and \textsc{CCI} output all passages possessing attribution scores above a threshold.
The threshold is zero for \textsc{AttnUnion} and \textsc{CCI}, and is 0.55 for \textsc{HSSAvg}.
The other hyperparameters of these fine-grained attribution methods are the same as the previous experiments.

For sentence-level methods, self-citation generates with 2-shot, temperature of 1.0, and top-$p$ of 0.95; \textsc{AttrFirst} generates with 1-shot for content selection, 4-shot for fusion in context, temperature of 0.3, max retry number of 5.

\noindent\textbf{Results.} We repeat all experiments three times with different seeds (except \textsc{AttrFirst} and \textsc{CCI} due to their heavy computational overload) and take the average results.
\textsc{AttrFirst} failed on 355 and 448 instances on ELI5 and ASQA, respectively, and we report the results measured on the successful instances.
The generation quality and citation quality results are shown in Table~\ref{tab:sentence-level-attribution} and Table~\ref{tab:generation-quality}, respectively.
As the results show, \textsc{AttnUnion} consistently outperforms other fine-grained attributors and improves the citation quality of \textsc{SelfCitation} and \textsc{AttrFirst}, suggesting its application for improving various attributors.

\subsection{Attribution Latency}
\label{sec:eval-latency}

% \usepackage{booktabs}
\begin{table}[t]
\centering
\resizebox{\linewidth}{!}{
\begin{tabular}{@{}lccc@{}}
\toprule
\textbf{}    & \textbf{QuoteSum} & \textbf{VERI-GRAN} & \textbf{ASQA} \\
\midrule
\rowcolor{gray!20}\multicolumn{4}{c}{\textbf{Baselines}}        \\
%\textsc{SelfCitation} & -                & -                 & 1962.7 \\ 
%\textsc{AttrFirst}    & -                 & -                 & 61706.0$_{(4)}$ \\
\textsc{CCI}          & 2921.8 & 7213.7$_{(3)}$ & 21780.5$_{(3)}$ \\
\textsc{HSSAvg}       & 84.5   & 538.5          & \underline{184.4} \\
\textsc{AttnUnion}$^\clubsuit$ & 141.9 & 1679.5 & 790.5 \\
\rowcolor{gray!20}\multicolumn{4}{c}{\textbf{Our Methods}}      \\
\textsc{AttnUnion}    & \textbf{22.7}     & \textbf{265.0}    & \textbf{123.4} \\
\textsc{AttnUnionDep} & \underline{54.0}  & \underline{427.4} & - \\
\bottomrule
\end{tabular}
}
\caption{Average Latency per target span (ms) with backbone of Qwen2 7B. By default, we ran this experiment on a single GPU, except for entries that met the OOM problem. The OOM entries were run on multiple GPUs, marked by a subscript indicating the number of GPUs used. The best entry is marked in bold and the second best entry is underlined. The Hugginface implementation of \textsc{AttnUnion} is marked by $\clubsuit$.}
\label{tab:latency}
\end{table}

\noindent\textbf{Research Question.} From a practical point of view, we ask: \textit{is our method faster than previous works?}

\noindent\textbf{Experiment Setting.} We compare our method with all fine-grained attribution baselines and \textsc{AttnUnion} based on Huggingface implementation of calculating attention weights.
The experiments are conducted on QuoteSum, VERI-GRAN, and ASQA with the backbone of an NF4-quantized Qwen2 7B and the device of a single NVIDIA RTX 3090 24GB GPU.
\textsc{HSSAvg} uses the same early exit as \textsc{AttnUnion} for a fair comparison.
We input instances of the datasets\footnote{On ASQA, the response is the generation of self-citation with citations removed.} one by one to the methods. 
The latency is measured by the average time consumed per target span.
% and these latencies do not include the latency of self-citation.
%, as the latency of \textsc{AttrFirst} also includes the generation latency.

\noindent\textbf{Results.} The results are shown in Table~\ref{tab:latency}.
Our implementation of \textsc{AttnUnion} largely outperforms other methods, with an average latency of 265.0 ms on the long-context dataset VERI-GRAN, and \textsc{AttnUnionDep} is the second.
%, demonstrating our methods' potential to be applied in real-time attribution systems.

\noindent\textbf{Insights.} Considering the early exits of \textsc{AttnUnion} and \textsc{HSSAvg} are the same, \textsc{AttnUnion} should have a similar latency to \textsc{HSSAvg}.
However, the latency of \textsc{AttnUnion} is much less than \textsc{HSSAvg}.
This is because the \textsc{AttnUnion} does not need to recompute average hidden states for each new target token as \textsc{HSSAvg}; instead, \textsc{AttnUnion} reuses token-wise attribution for all target spans in the same response.

% \section{Introduction}

% These instructions are for authors submitting papers to *ACL conferences using \LaTeX. They are not self-contained. All authors must follow the general instructions for *ACL proceedings,\footnote{\url{http://acl-org.github.io/ACLPUB/formatting.html}} and this document contains additional instructions for the \LaTeX{} style files.

% \section{Engines}

% To produce a PDF file, pdf\LaTeX{} is strongly recommended (over original \LaTeX{} plus dvips+ps2pdf or dvipdf). Xe\LaTeX{} also produces PDF files, and is especially suitable for text in non-Latin scripts.

% By default, the box containing the title and author names is set to the minimum of 5 cm. % If you need more space, include the following in the preamble:
% \begin{quote}
% \begin{verbatim}
% \setlength\titlebox{<dim>}
% \end{verbatim}
% \end{quote}
% where \verb|<dim>| is replaced with a length. Do not set this length smaller than 5 cm.

%\subsection{Citations}

%You can use the command \verb|\citealp| (alternative cite without parentheses) to get ``author, year'' citations, which is useful for using citations within parentheses (e.g. \citealp{Gusfield:97}).

\section{Conclusion}

This work proposes a novel fine-grained attribution method, leveraging attention weights and dependency parsing. 
The experiments show that our method is the new SOTA fine-grained attributor and generalizes well to sentence-level attribution.
Moreover, our method is much faster than previous works, showing its potential to be applied in real-time attribution systems.

\section{Limitations}

A limitation of our evaluation is that the target spans of QuoteSum and VERI-GRAN are selected by models, which might not reflect the real application scenario where target spans are selected by human users.
This limitation could be resolved by collecting user-selected target spans in the future. 
Another limitation is that QuoteSum and VERI-GRAN benchmarks focus on attributing verbatim spans from the documents.
Although \citet{phukan2024peering} observe that LLMs tend to produce verbatim answers, it is interesting to evaluate attribution with more abstractive answers in future work.
In addition, in dependency parsing augmentation, the expansion on the dependency parse tree is rule-based (e.g., recognizing coordinating components by relation "conj"), which could be improved by machine learning in the future.
Finally, our work only considers attribution in English.
Although the attention-based method can be easily adapted to LLMs of other languages, it may take effort (but not difficult, as we show in Appendix \ref{sec:chinese-experiment}) to adapt our dependency parsing augmentation to other languages since dependency parsing is language-specific.

%\section*{Acknowledgments}

\bibliography{custom}

\appendix

% \usepackage{booktabs}
\begin{table*}[t]
\resizebox{\linewidth}{!}{
\begin{tabular}{@{}lcccccccc@{}}
\toprule
\textbf{} & \textbf{\begin{tabular}[c]{@{}c@{}}Target\\ Granularity\end{tabular}} & \textbf{\begin{tabular}[c]{@{}c@{}}Evidence\\ Granularity\end{tabular}} & \textbf{\#instance} & \textbf{\begin{tabular}[c]{@{}c@{}}\#target\\ per instance\end{tabular}} & \textbf{\begin{tabular}[c]{@{}c@{}}\#evidence\\ per instance\end{tabular}} & \textbf{\begin{tabular}[c]{@{}c@{}}Document\\ Len.\end{tabular}} & \textbf{\begin{tabular}[c]{@{}c@{}}Question\\ Len.\end{tabular}} & \textbf{\begin{tabular}[c]{@{}c@{}}Answer\\ Len.\end{tabular}} \\
\midrule
\rowcolor{gray!20}\multicolumn{9}{c}{\textit{Fine-grained Attribution}} \\
QuoteSum  & span & passage & 1319 & 4.6 & 3.4 & 1958.1 & 43.6 & 275.4 \\
VERI-GRAN & span & sentence & 197 & 1.2 & 69.1 & 8211.3 & 52.9 & 398.9 \\
\rowcolor{gray!20}\multicolumn{9}{c}{\textit{Sentence-level Attribution}} \\
ELI5      & sentence & passage & 1000 & 6.7 & 5.0 & 2906.3 & 92.3 & 699.2 \\
ASQA      & sentence & passage & 948 & 3.5 & 5.0 & 3005.6 & 47.2 & 427.0 \\
\bottomrule
\end{tabular}
}
\caption{Statistics of datasets used in experiments, where lengths are measured by the numbers of characters. 
% QuoteSum and VERI-GRAN are fine-grained attribution datasets, and ELI5 is a sentence-level attribution dataset.
}
\label{tab:statistics}
\end{table*}

\begin{table}[t]
\centering
\resizebox{\linewidth}{!}{
\begin{tabular}{lccccc}
\toprule
& \multicolumn{2}{c}{\textbf{ELI5}}    & \multicolumn{2}{c}{\textbf{ASQA}} \\
\cmidrule(r){2-3} \cmidrule(r){4-5}
& \textbf{MAUVE} & \textbf{Claim Rec.} & \textbf{MAUVE} & \textbf{EM Rec.} \\
\midrule
\rowcolor{gray!20}\multicolumn{5}{c}{\textit{Qwen2}} \\
\textsc{SelfCitation} & 23.5 & 12.4 & 80.5 & 39.6 \\
\textsc{AttrFirst}    & 55.0 & 4.5 & 55.9 & 35.2 \\
\rowcolor{gray!20}\multicolumn{5}{c}{\textit{Llama2}} \\
\textsc{SelfCitation} & 32.9 & 12.3 & 56.9 & 30.1 \\
\bottomrule
\end{tabular}
}
\caption{The generation quality results of \textsc{SelfCitation} and \textsc{AttrFirst}.}
\label{tab:generation-quality}
\end{table}

\section{Pseudo-code for \textsc{AttnUnion}}

The pseudo-code for \textsc{AttnUnion} is Algorithm \ref{alg:basic-algorithm}.

\begin{algorithm}[t]
\caption{The Basic Algorithm}
\label{alg:basic-algorithm}
\begin{algorithmic}
\State {\bfseries Input:} Document tokens $\mathbf{d} = (d_1, ..., d_c)$, question tokens $\mathbf{q} = (q_1, ..., q_m)$, Response tokens $\mathbf{r} = (r_1, ..., r_n)$, the target span $\mathbf{t}\subset \mathbf{r}$, and the model $\mathcal{M}$ that outputs similarity metric.
\State {\bfseries Output:} The evidence tokens of the target span $T$ and their scores.
    \State $\mathbf{S} \leftarrow \mathcal{M}(\mathbf{d+q+r}) \in \mathbb{R}^{r\times (c+m)}$
    %\State $\mathbf{W} \leftarrow \frac 1H \sum_{h=1}^H \mathbf{A}^{(L^*, h)} \in \mathbb{R}^{r\times c}$
    \State $w \leftarrow \text{dictionary}()$ \Comment{The attribution scores}
    \For{$r_i \in \mathbf{t}$} \Comment{Aggregate evidence and scores}
        \For{$j\in \text{range}(\mathbf{d})\cap\arg\text{top-}k(\mathbf{S}_i)$}
            \If{$j \in w.keys()$}
                \State $w[j] \leftarrow w[j]+\mathbf{S}_{ij}$
            \Else
                \State $w[j] \leftarrow \mathbf{S}_{ij}$
            \EndIf
        \EndFor
    \EndFor
    \State $new\_w \leftarrow \text{dictionary}()$
    \For{$i \in w.keys()$} \Comment{Remove isolated tokens}
        \State $isolated \leftarrow True$
        \For{$j \textbf{ in } i-\tau \textbf{ to } i+\tau$}
            \If{$j\neq i$ \textbf{and} $j\in w.keys()$}
                \State $isolated \leftarrow False$
                \State \textbf{break}
            \EndIf
        \EndFor
        \If{$\neg isolated$}
            \State $new\_w[i] \leftarrow w[i]$
        \EndIf
    \EndFor
    \State {\bfseries return} $new\_w$
\end{algorithmic}
\end{algorithm}

\section{Word-Token Alignment in \textsc{Dep}}
\label{sec:word-token-alignment}

With words and tokens distinguished, the atomic-fact recognition method $\mathcal{A}$ should be updated to 
$$
\mathcal{A}(r_i)\leftarrow\psi\left(\bigcup_{w\in\phi(r_i)}\widetilde{\mathcal{A}}(w))\right),
$$
where $\widetilde{\mathcal{A}}$ is the atomic fact recognition method introduced in Sec.~\ref{sec:dependency-parsing-augmentation}, $\phi$ is the token-to-words mapping, and $\psi$ is words-to-tokens mapping.
The $\phi$ maps a token to the minimum set of words that cover the token.
The $\psi$ maps a set of words to the minimum set of tokens that covers these words.

\section{Details of Excluding Irrelevant Coordinating Constituents}
\label{sec:excluding-coordinates}

\begin{figure}
    \centering
    \includegraphics[width=\linewidth]{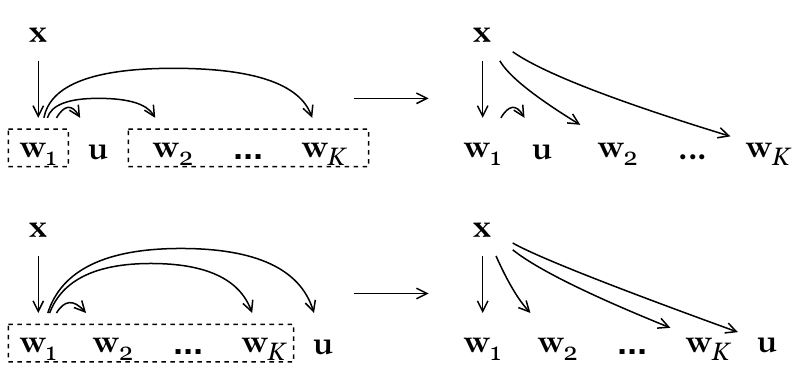}
    \caption{An illustration of reforming the coordinate structures, where the words framed by dash lines are coordinate components.}
    \label{fig:reform-tree}
\end{figure}

This section describes how to exclude irrelevant coordinating constituents, given that $\mathcal{A}(r_i)$ has included all successors of $v$ (the closest verb ancestor of $r_i$).
In a dependency parse tree, a \textit{coordinating structure} is multiple words $\mathbf{w}_1, \mathbf{w}_1, ..., \mathbf{w}_K, K > 1$ that satisfy
\begin{equation}
    \begin{cases}
        \text{label}(\mathbf{w_1}) = ... = \text{label}(\mathbf{w}_{K-1}), \\
        \text{label}(\mathbf{w}_K)=\text{``conj''}, \\
        \text{head}(\mathbf{w}_2) = ... = \text{head}(\mathbf{w}_K) = \mathbf{w}_1,
    \end{cases}
\end{equation}
where head($\mathbf{w}_i$) represent the parent of $\mathbf{w}_i$, and label($\mathbf{w}_i$) represent the corresponding relation type between the parent and $\mathbf{w}_i$.
Considering that the first component is special, we name the first component as the \textit{leader} of the coordinating structure.
The algorithm is outlined as follows.

\noindent\textbf{Input.} $r_i$: the word to augment; $v$: the closest verb ancestor of $r_i$; $\mathcal{A}(r_i)$: the atomic fact elements that have collected $v$ and its successors; the heads and labels of all words.

\noindent\textbf{Output.} The $\mathcal{A}(r_i)$ with irrelevant coordinate components removed.

\noindent\textbf{1. Identify coordinating structures.} The coordinating structures are searched by enumerating the potential leaders (shown as follows).

\begin{lstlisting}[   % 进行参数设置
 language=Python, % 设置语言
 basicstyle=\small\ttfamily, % 设置字体族
 breaklines=true, % 自动换行
 keywordstyle=\bfseries\color{blue},
 emph={find_coordinations},
 emphstyle={\bfseries\color{yellow!60!black}},
 commentstyle=\itshape\color{green!50!black},
 stringstyle=\bfseries\color{orange!70!black},
 columns=flexible
]  
def find_coordinations(
    dep_head, 
    dep_label
):
    coordinations = []
    in_coordination_words = set()
    for j in range(len(dep_head)-1):
        if j in in_coordination_words:
            continue
        new_coordination = [j]
        for k in range(j+1, len(dep_head)):
            case1 = get_head(dep_head, k) == j and dep_label[k] == dep_label[j]
            case2 = get_head(dep_head, k) == j and dep_label[k] == 'conj'
            if case1 or case2:
                new_coordination.append(k)

        if len(new_coordination) > 1:
            coordinations.append(sorted(new_coordination))
            in_coordination_words.update(new_coordination)

    return coordinations
\end{lstlisting}

\noindent\textbf{2. Reform the tree.} For the convenience of the following process, we temporarily reform the local structures for all coordinate structures, as Fig.~\ref{fig:reform-tree} shows.
We replace the heads of non-leader components with the head of the leader, ending the asymmetric relationship between the leader and the other components.
For other children of the leader, we retain its head if it precedes the first non-leader component (as the upper half of Fig. \ref{fig:reform-tree} shows); otherwise, we replace its head with the head of the leader (as the lower half of Fig. \ref{fig:reform-tree} shows).

\noindent\textbf{3. Identify the path from $v$ to $r_i$.}

\noindent\textbf{4. Process Coordinating Structures that intersect with $v\rightarrow r_i$.} For coordinating structures that intersect with $v\rightarrow r_i$, the tree retains the intersection and removes all other components. 

\noindent\textbf{5. Process Coordinating Structures that do not intersect with $v\rightarrow r_i$.} For non-intersecting coordinating structures $\mathcal{G}$, the algorithm first determines whether there is a parallel coordinating structure of it, where parallel coordinating structures are those that have the same number of constituents, e.g., (``one million dollars,'' ``two million dollars'') and (``2012,'' ``2013'').
If so, denoting the parallel coordinating structure as $\mathcal{G}'$ and its $i$-th component is retained, then the algorithm deletes all components of $\mathcal{G}$ except the $i$-th component from the dependency parse tree.

\noindent\textbf{6. Recollect.} The algorithm recollects all $v$'s successors (except punctuation marks) with the new tree, yielding the final $\mathcal{A}(r_i)$.

\section{Details of Huggingface Implementation of \textsc{AttnUnion}}
\label{sec:huggingface-implementation}

\begin{lstlisting}[   % 进行参数设置
 language=Python, % 设置语言
 basicstyle=\small\ttfamily, % 设置字体族
 breaklines=true, % 自动换行
 keywordstyle=\bfseries\color{blue},
 emph={forward, forward_stage1, forward_stage2},
 emphstyle={\bfseries\color{yellow!60!black}},
 commentstyle=\itshape\color{green!50!black},
 stringstyle=\bfseries\color{orange!70!black},
 columns=flexible
]  
def forward_stage1(self, batch):
    """stage 1: do not output attention, output kv cache on prompt_ids
    """
    outputs = self.model(
        input_ids=batch['prompt_ids'][:,:-1].to(self.device),  # leave the last prompt token to forward in stage2
        attention_mask=batch['prompt_mask'][:,:-1].to(self.device),
        output_attentions=False, 
        use_cache=True,
        return_dict=True, 
    )
    return outputs.past_key_values

def forward_stage2(self, batch, past_key_values):
    """stage2: output response-to-prompt attentions
    """
    attention_mask = torch.cat([batch['prompt_mask'][:,:-1], batch['response_mask']], dim=1).to(self.device)
    input_ids = torch.cat([batch['prompt_ids'][:,-1:], batch['response_ids'][:,:-1]], dim=1).to(self.device)
    outputs = self.model(
        input_ids=input_ids,
        attention_mask=attention_mask,
        past_key_values=past_key_values,
        output_attentions=True, 
        use_cache=True,
        return_dict=True, 
    )
    return outputs.attentions

def forward(self, batch):
    """output response-to-prompt attentions for causal LMs
    """
    with torch.no_grad():
        past_key_values = self.forward_stage1(batch)
        attentions = self.forward_stage2(batch, past_key_values)
        attentions = attentions[self.output_attentions_layer]
    return attentions
\end{lstlisting}

Using Huggingface Transformers to output self-attention weights on the concatenation of the prompt and the response is impractical since this approach will incur tremendous memory overload of $\mathcal{O}((c+m+n)^2)$.

A more memory-efficient approach is to use the response as the query and the concatenation of prompt and response as the key, which incurs a memory overload of $\mathcal{O}(n\times(c+m))$ and requires the KV cache.
We choose the latter memory-efficient approach as our baseline in Sec.~\ref{sec:eval-latency}, which is illustrated by the \textcolor{yellow!60!black}{\texttt{forward}} method in the code above.

\section{Statistics of Datasets}

The statistics of datasets are shown in Table~\ref{tab:statistics}.

%\section{Results of Ablating Dependency Parsing on Qwen2 7B}

%\begin{figure}[t]
%    \centering
%    \includegraphics[width=\linewidth]{figs/qwen2_ablating_dep.pdf}
%    \caption{Results of ablating dependency parsing on Qwen2 7B. Here AU and HU represent \textsc{AttnUnion} and \textsc{HSSUnion}, respectively.}
%    \label{fig:qwen2-ablating-dep}
%\end{figure}

\section{Experiments on \textsc{AttnUnionDep}'s Sensitivity to Hyperparameters}
\label{sec:hyperparameter-sensitivity}

Here, we consider the following hyperparameters:

1. $L^*$: the layer to extract attention weights;

2. $k$: how many top-scored prompt tokens are selected as evidence for each response token;

3. $\tau$: the threshold for recognizing isolated tokens.

\begin{figure}
    \centering
    \includegraphics[width=0.95\linewidth]{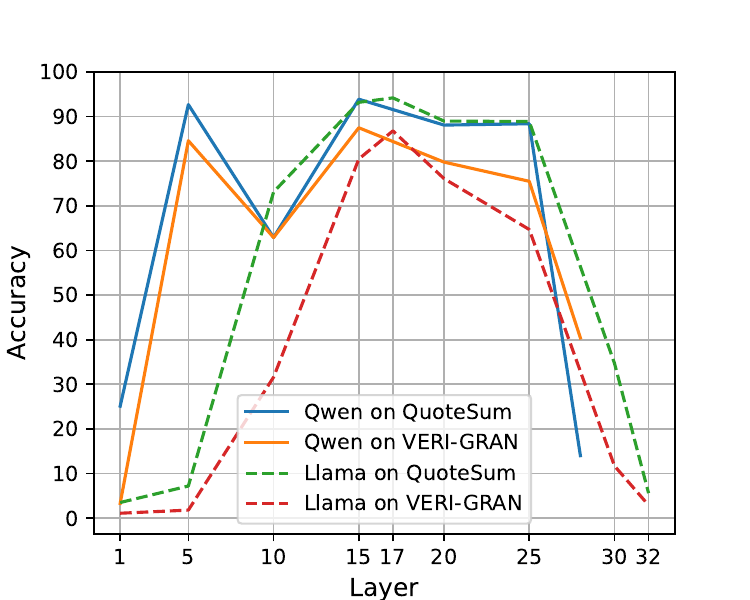}
    \caption{The validation accuracy of \textsc{AttnUnionDep} against the layer from which the attention weights are extracted (fixing $k = 2, \tau = 2$). For Qwen2 7B, $L = 28$, and $\lfloor L/2\rfloor +1=15$. For Llama2 7B, $L = 32$, , and $\lfloor L/2\rfloor +1 = 17$.}
    \label{fig:tuning-layer}
\end{figure}

\begin{figure}
    \centering
    \includegraphics[width=0.95\linewidth]{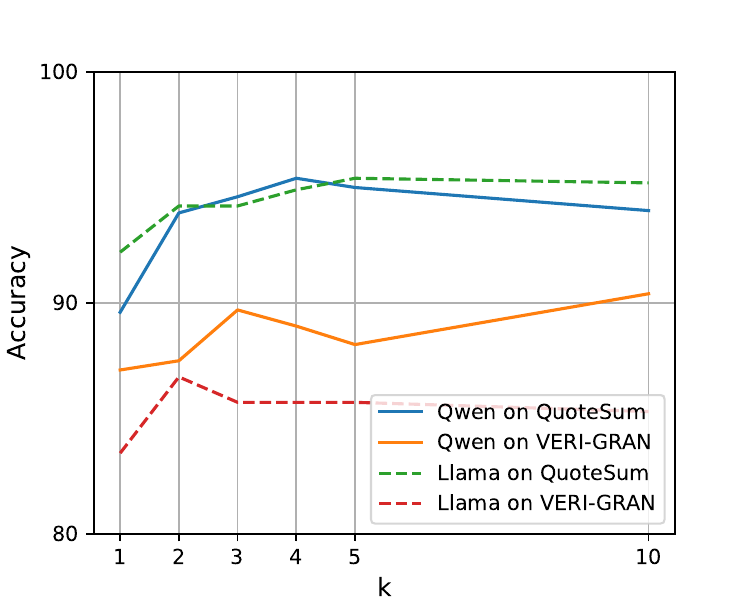}
    \caption{The validation accuracy of \textsc{AttnUnionDep} against $k$ (fixing $\tau = 2$, $L^* = 15$ and 17 for Qwen2 7B and Llama2 7B, respectively).}
    \label{fig:tuning-k}
\end{figure}

\begin{figure}
    \centering
    \includegraphics[width=0.95\linewidth]{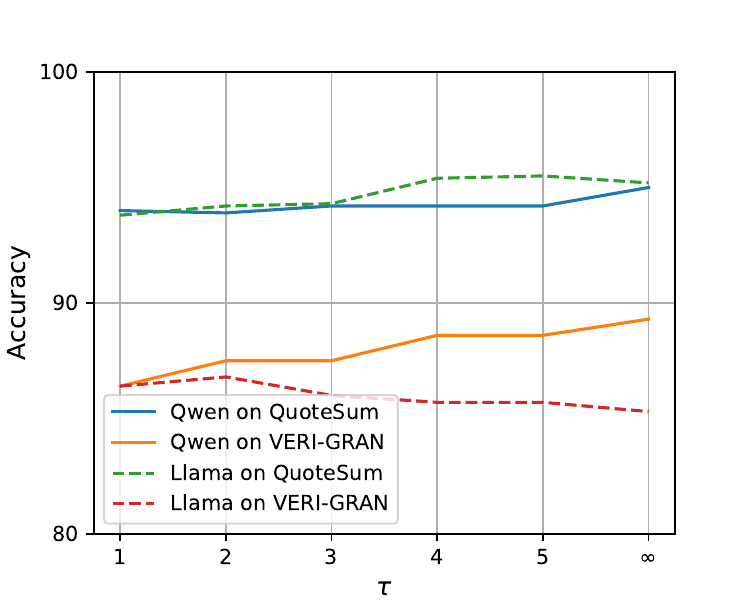}
    \caption{The validation accuracy of \textsc{AttnUnionDep} against $\tau$ (fixing $k = 2$, $L^* = 15$ and 17 for Qwen2 7B and Llama2 7B, respectively). Here, $\tau = \infty$ means no filtering out isolated evidence tokens.}
    \label{fig:tuning-tau}
\end{figure}

The experiments are conducted on validation sets of QuoteSum and VERI-GRAN, with the metric of accuracy (\%), the attributor of \textsc{AttnUnionDep}, and the models of Qwen2 7B and Llama 7B. The results are shown in Fig.~\ref{fig:tuning-layer}, \ref{fig:tuning-k}, and \ref{fig:tuning-tau}. The performance is relatively stable across a range of hyperparameters. Note that our choice of hyperparameters may not be optimal, because we did not extensively tune them but qualitatively chose them by human evaluation. We found under these hyperparameters, the proposed method provides the most human-friendly (neat and fragmentless) attribution.

%\begin{table}[ht]
%    \centering
%    \resizebox{\linewidth}{!}{
%    \begin{tabular}{ccccccccc}
%    \hline
%     & & 1 & 5 & 10 & 15 & 20 & 25 & 28 \\ \hline
%    \multirow{2}{*}{\texttt{Qw.}}& \texttt{QS} & 25.1 & 92.7 & 62.9 & 93.9 & 88.1 & 88.4 & 14.0 \\ \cmidrule{2-9}
%                     & \texttt{VG} & 3.3 & 84.6 & 62.9 & 87.5 & 79.8 & 75.5 & 40.4 \\ \hline
%    \multirow{2}{*}{\texttt{Ll.}} & \texttt{QS} & & & & & & & \\ \cmidrule{2-9}
%    & \texttt{VG} & & & & & & & \\ \hline
%    \end{tabular}
%    }
%    \caption{}
%    \label{tab:tuning-layer}
%\end{table}

\section{Generation Qualities of \textsc{SelfCitatoin} and \textsc{AttrFirst}}

The generation qualities of \textsc{SelfCitation} and \textsc{AttrFirst} are shown in Table~\ref{tab:generation-quality}.

\section{Evaluating Attribution on Chinese Synthetic Datasets}
\label{sec:chinese-experiment}

We conducted experiments on Chinese synthetic datasets to show that our method can be migrated to languages other than English.
The synthetic Chinese datasets are constructed by translating the answers from QuoteSum and VERI-GRAN to Chinese and maintaining the questions and the passages in English. Specifically, each answer is processed as follows (during this process, the annotations of the target spans are maintained):

1. segmenting the answer by the boundaries of target spans;

2. separately translating each segment into Chinese;

3. concatenating all translated segments to build the translated answer.

We adapt the dependency parsing augmentation to Chinese without modifying the rules (except for mapping Chinese dependency parsing labels to the English counterpart). We compare \textsc{HSSAvg}, \textsc{AttnUnion}, and \textsc{AttnUnionDep} using the backbone of Qwen2 7B. For all methods, we use the same hyperparameters as in Sec. \ref{sec:eval-fine-grained-attribution}. The results are shown in Table \ref{tab:chinese-results}, which shows that our method \textsc{AttnUnionDep} still outperforms \textsc{HSSAvg} and \textsc{AttnUnion}.

\begin{table}[ht]
\centering
\resizebox{\linewidth}{!}{
\begin{tabular}{ccc}
\toprule
    \multirow{2}{*}{\textbf{Method}} & \textbf{Translated} & \textbf{Translated} \\ 
    & \textbf{QuoteSum} & \textbf{VERI-GRAN} \\ \midrule
    \textsc{HSSAvg} & 76.1 & 61.6 \\ 
    \textsc{AttnUnion} & 76.0 & 65.3 \\ 
    \textsc{AttnUnionDep} & \textbf{87.0} & \textbf{72.5} \\ \bottomrule
\end{tabular}
}
\caption{Accuracy (\%) on the synthetic Chinese datasets, where the best entries are marked in bold.}
\label{tab:chinese-results}
\end{table}

\section{An Additional Alternative of \textsc{Dep}}
\label{sec:additional-dep-alternative}

During the review of this paper, a reviewer proposed an interesting alternative of \textsc{Dep}: considering the self-attention among response tokens, the latter response tokens can be attributed to previous response tokens and this might also reveal some relationship between them (as what \textsc{Dep} does). For instance, ``2012'' might be attributed to ``one million dollars''. Then the attribution of ``one million dollars'' could be updated with the attribution scores from ``2012''.

We implemented the attribution among the response tokens using the same hyperparameters (except $\tau$; no filtering out isolated evidence tokens in this attribution) as the attribution between response and prompt tokens.
The results are shown in Table \ref{tab:additional-dep-alternative}. Here we name the alternative \textsc{AugmentByAttn}. We also evaluate a variant of \textsc{AugmentByAttn} that limits the augmentation tokens in the target span's local sentence, to ensure the augmentation tokens are more relevant to the target span. As the results show, \textsc{AugmentByAttn} and its variant improve \textsc{AttnUnion} but are not as good as dependency parsing augmentation, indicating dependency parsing augmentation is more accurate in recognizing semantically relevant tokens to the target span.

\begin{table}[ht]
\centering
\resizebox{\linewidth}{!}{
\begin{tabular}{lcc}
\toprule
    ~ & \textbf{QuoteSum} & \textbf{VERI-GRAN} \\ \midrule
    \textsc{AttnUnion} & 79.4 & 70.9 \\ 
    \textsc{AugmentByAttn} & 86.0 & 78.4 \\ 
    \textsc{AugmentByAttn} variant & 86.2 & 77.8 \\ 
    \textsc{AttnUnionDep} & \textbf{93.3} & \textbf{84.6} \\ \bottomrule
\end{tabular}
}
\caption{Evaluation results for addition alternatives of \textsc{Dep}, i.e., \textsc{AugmentByAttn} and its variant, where the backbone is Qwen2 7B.}
\label{tab:additional-dep-alternative}
\end{table}

\end{document}